\documentclass{article}

\usepackage[preprint,nonatbib]{neurips_2020}
\usepackage[utf8]{inputenc} 
\usepackage[T1]{fontenc}    
\usepackage{hyperref}       
\usepackage{url}            
\usepackage{booktabs}       
\usepackage{amsfonts}       
\usepackage{nicefrac}       
\usepackage{microtype}      

\usepackage{hyperref}
\hypersetup{
    colorlinks=true,
    linkcolor=blue,
    filecolor=magenta,      
    urlcolor=blue,
}

\usepackage{wrapfig}
\usepackage{graphicx}
\usepackage{caption}
\usepackage{subcaption}
\usepackage{floatrow}
\usepackage{booktabs}
\usepackage{varwidth}
\usepackage{todonotes}
\usepackage[capitalize]{cleveref}

\newfloatcommand{capbtabbox}{table}[][\FBwidth]
\usepackage{blindtext}

\title{ChemBERTa: Large-Scale Self-Supervised Pretraining for Molecular Property Prediction}

\author{
  Seyone Chithrananda \\
  University of Toronto\\
  \texttt{seyone.chithrananda@utoronto.ca} \\
  \And 
  Gabriel Grand \\
  Reverie Labs \\
  \texttt{gabe@reverielabs.com} \\
  \And
  Bharath Ramsundar \\
  DeepChem \\
  \texttt{bharath.ramsundar@gmail.com}
}

\begin{document}

\maketitle

\begin{abstract}
GNNs and chemical fingerprints are the predominant approaches to representing molecules for property prediction. However, in NLP, transformers have become the de-facto standard for representation learning thanks to their strong downstream task transfer. In parallel, the software ecosystem around transformers is maturing rapidly, with libraries like HuggingFace and BertViz enabling streamlined training and introspection. In this work, we make one of the first attempts to systematically evaluate transformers on molecular property prediction tasks via our ChemBERTa model. While not at state-of-the-art, ChemBERTa scales well with pretraining dataset size, offering competitive downstream performance on MoleculeNet and useful attention-based visualization modalities. Our results suggest that transformers offer a promising avenue of future work for molecular representation learning and property prediction. To facilitate these efforts, we release a curated dataset of 77M SMILES from PubChem suitable for large-scale self-supervised pretraining. 
\end{abstract} 

\section{Motivation}

Molecular property prediction has seen a recent resurgence thanks to the success of graph neural networks (GNNs) on various benchmark tasks \cite{duvenaud2015convolutional, kearnes2016molecular, kipf2016semi, gilmer2017neural, coley2017convolutional, Chemprop}. However, data scarcity remains a fundamental challenge for supervised learning in a domain in which each new labelled data point requires costly and time-consuming laboratory testing. Determining effective methods to make use of large amounts of unlabeled structure data remains an important unsolved challenge.

Over the past two years, the transformer \cite{attention, bert} has emerged as a robust architecture for learning self-supervised representations of text. Transformer pretraining plus task-specific finetuning provides substantial gains over previous approaches to many tasks in natural language processing (NLP) \cite{lan2019albert, electra, raffel2020exploring}. Meanwhile, software infrastructure for transformers is maturing rapidly: HuggingFace \cite{huggingface} provides streamlined pretraining and finetuning pipelines, while packages like BertViz \cite{bertviz} offer sophisticated interfaces for attention visualization. Given the availability of millions of SMILES strings, transformers offer an interesting alternative to both expert-crafted and GNN-learned fingerprints. In particular, the masked language-modeling (MLM) pretraining task \cite{bert} commonly used for BERT-style architectures is analogous to atom masking tasks used in graph settings \cite{hu2019strategies}. Moreover, since modern transformers are engineered to scale to massive NLP corpora, they offer practical advantages over GNNs in terms of efficiency and throughput.

Though simple in concept, the application of transformers to molecular data presents several questions that are severely underexplored. For instance: How does pretraining dataset size affect downstream task performance? What tokenization strategies work best for SMILES? Does replacing SMILES with a more robust string representation like SELFIES \cite{selfies} improve performance? We aim to address these questions via one of the first systematic evaluations of transformers on molecular property prediction tasks.

\section{Related Work}

In cheminformatics, there is a long tradition of training language models directly on SMILES to learn continuous latent representations \cite{xu2017seq2seq, kusner2017grammar, gomez2018automatic}. Typically, these are RNN sequence-to-sequence models and their goal is to facilitate auxiliary lead optimization tasks; e.g., focused library generation \cite{segler2018generating}. Thus far, discussion of the transformer architecture in chemistry has been largely focused on a particular application to reaction prediction \cite{schwaller2019molecular}. 

Some recent work has pretrained transformers for molecular property prediction and reported promising results \cite{honda2019smiles, maziarka2020molecule}. However, the datasets used for pretraining have been relatively small (861K compounds from ChEMBL and 2M compounds from ZINC, respectively). Other work has used larger pretraining datasets (18.7M compounds from ZINC) \cite{wang2019smiles} but the effects of pretraining dataset size, tokenizer, and string representation were not explored. In still other work, transformers were used for supervised learning directly without pretraining \cite{chen2019path}. 

Recently, a systematic study of self-supervised pretraining strategies for GNNs helped to clarify the landscape of those methods \cite{hu2019strategies}. Our goal is to undertake a similar investigation for transformers to assess the viability of this architecture for property prediction.

\section{Methods}


ChemBERTa is based on the RoBERTa \cite{roberta} transformer implementation in HuggingFace \cite{huggingface}. Our implementation of RoBERTa uses 12 attention heads and 6 layers, resulting in 72 distinct attention mechanisms. So far, we have released 15 pre-trained ChemBERTa models on the \href{https://huggingface.co/seyonec/ChemBERTa-zinc-base-v1}{Huggingface's model hub;} these models have collectively received over 30,000 Inference API calls to date.\footnote{The main model directory can be viewed \href{https://huggingface.co/seyonec}{here}. Each model includes the specific tokenizer (BPE, SMILES-tokenized), representation (SMILES, SELFIES) and number of training steps ('150k') appended in its name.} 

We used the popular Chemprop library for all baselines \cite{Chemprop}. We trained the directed Message Passing Neural Network (D-MPNN) with default hyperparameters as well as the sklearn-based \cite{pedregosa2011scikit} Random Forest (RF) and Support Vector Machine (SVM) models from Chemprop, which use 2048-bit Morgan fingerprints from RDKit \cite{ecfp, rdkit}.

\subsection{PreTraining on PubChem 77M}

We adopted our pretraining procedure from RoBERTa, which masks 15\% of the tokens in each input string. We used a max. vocab size of 52K tokens and max. sequence length of 512 tokens. We trained for 10 epochs on all PubChem subsets except for the 10M subset, on which we trained for 3 epochs to avoid observed overfitting. Our hypothesis is that, in learning to recover masked tokens, the model forms a representational topology of chemical space that should generalize to property prediction tasks.

For pretraining, we curated a dataset of 77M unique SMILES from PubChem \cite{pubchem}, the world's largest open-source collection of chemical structures. The SMILES were canonicalized and globally shuffled to facilitate large-scale pretraining. We divided this dataset into subsets of 100K, 250K, 1M, and 10M. Pretraining on the largest subset took approx. 48 hours on a single NVIDIA V100 GPU. We make this dataset \href{https://deepchemdata.s3-us-west-1.amazonaws.com/datasets/pubchem_10m.txt.zip}{publicly available} and leave pretraining on the full 77M set to future work.

\subsection{Finetuning on MoleculeNet}

We evaluated our models on several classification tasks from MoleculeNet \cite{wu2018moleculenet} selected to cover a range of dataset sizes (1.5K - 41.1K examples) and medicinal chemistry applications (brain penetrability, toxicity, and on-target inhibition). These included the BBBP, ClinTox, HIV, and Tox21 datasets. For datasets with multiple tasks, we selected a single representative task: the clinical toxicity (CT\_TOX) task from ClinTox and the p53 stress-response pathway activation (SR-p53) task from Tox21. For each dataset, we generated an 80/10/10 train/valid/test split using the scaffold splitter from DeepChem \cite{deepchem}. During finetuning, we appended a linear classification layer and backpropagated through the base model. We finetuned models for up to 25 epochs with early stopping on ROC-AUC. We release a tutorial \href{https://github.com/deepchem/deepchem/blob/master/examples/tutorials/22_Transfer_Learning_With_HuggingFace_tox21.ipynb}{in DeepChem} which allows users to go through loading a pre-trained ChemBERTa model, running masked prediction tasks, visualizing the attention of the model on several molecules, and fine-tuning the model on the Tox21 SR-p53 dataset.

\section{Results}



\begin{figure}[ht!]
    \centering
    \includegraphics[width=\textwidth]{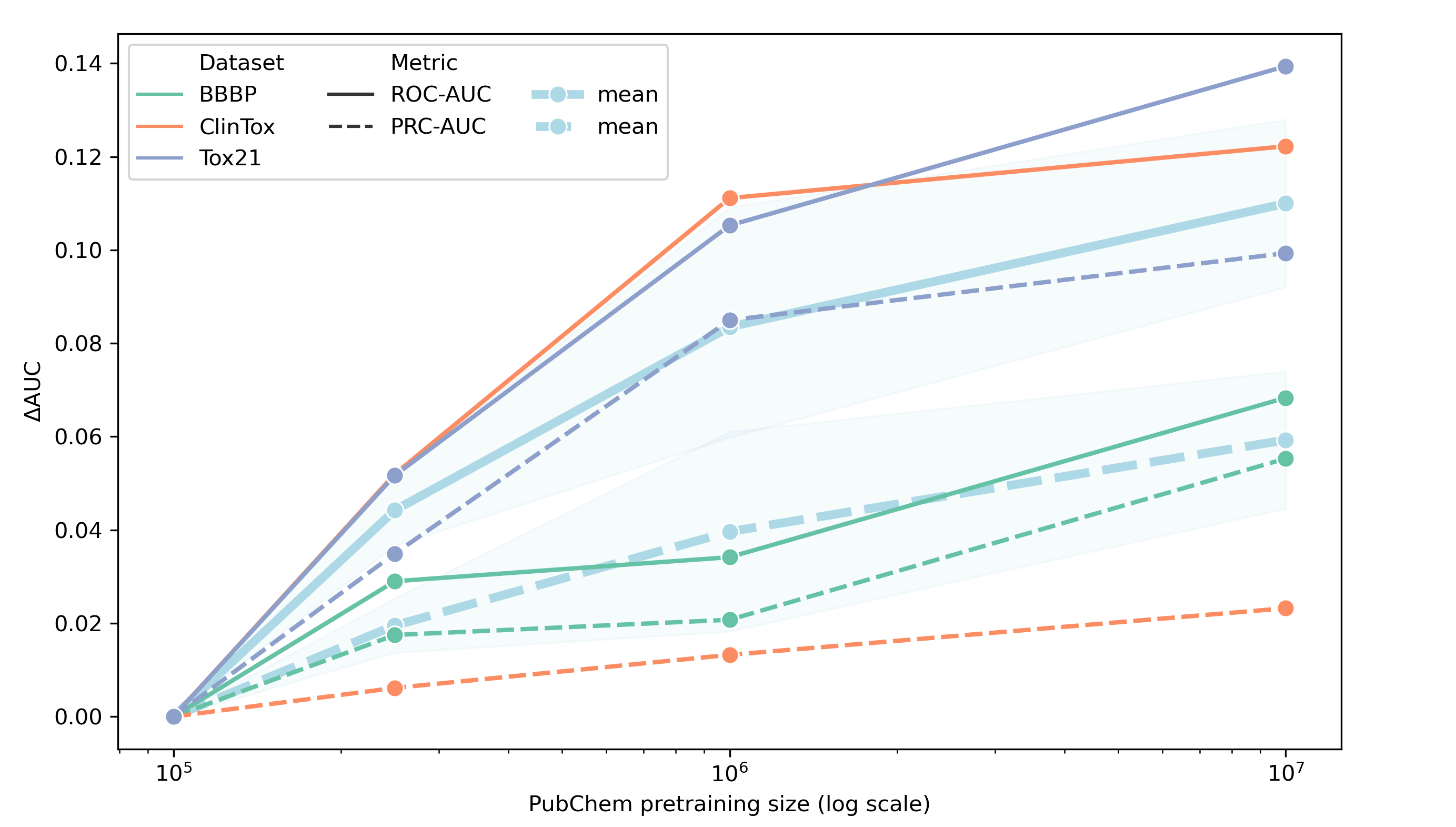}
    \caption{Scaling the pretraining size (100K, 250K, 1M, 10M) produces consistent improvements in downstream task performance on BBBP, ClinTox, and Tox21. Mean $\Delta$AUC across all three tasks with a 68\% confidence interval is shown in light blue.}
    \label{fig:scaling}
\end{figure}

On the MoleculeNet tasks that we evaluated, ChemBERTa approaches, but does not beat, the strong baselines from Chemprop  (\cref{tab:main-results}).\footnote{While Tox21 ROC-AUC is better than the baselines, PR-AUC is considerably lower.} Nevertheless, downstream performance of ChemBERTa scales well with more pretraining data (\cref{fig:scaling}). On average, scaling from 100K to 10M resulted in $\Delta \textrm{ROC-AUC} = +0.110$ and $\Delta \textrm{PRC-AUC} = +0.059$. (HIV was omitted from this analysis due to resource constraints.) These results suggest that ChemBERTa learns more robust representations with additional data and is able to leverage this information when learning downstream tasks.

\begin{table}[t]
\begin{tabular}{@{}lllllllll@{}}
\toprule
 & \multicolumn{2}{c}{\textbf{BBBP}} & \multicolumn{2}{c}{\textbf{ClinTox (CT\_TOX)}} & \multicolumn{2}{c}{\textbf{HIV}} & \multicolumn{2}{c}{\textbf{Tox21 (SR-p53)}} \\ 
 & \multicolumn{2}{c}{2,039} & \multicolumn{2}{c}{1,478} & \multicolumn{2}{c}{41,127} & \multicolumn{2}{c}{7,831} \\
 \midrule
 & ROC & PRC & ROC & PRC & ROC & PRC & ROC & PRC \\ \midrule
ChemBERTa 10M & 0.643 & 0.620 & 0.733 & 0.975 & 0.622 & 0.119 & \textbf{0.728} & 0.207 \\
D-MPNN & \textbf{0.708} & 0.697 & \textbf{0.906} & \textbf{0.993} & 0.752 & 0.152 & 0.688 & \textbf{0.429} \\
RF & 0.681 & 0.692 & 0.693 & 0.968 & \textbf{0.780} & \textbf{0.383} & 0.724 & 0.335 \\
SVM & 0.702 & \textbf{0.724} & 0.833 & 0.986 & 0.763 & 0.364 & 0.708 & 0.345 \\ \bottomrule
\end{tabular}
\caption{\label{tab:main-results}Comparison of ChemBERTa pretrained on 10M PubChem compounds and Chemprop baselines on selected MoleculeNet tasks. We report both ROC-AUC and PRC-AUC to give a full picture of performance on class-imbalanced tasks.
}
\end{table}

\vspace{-1em}


\subsection{Tokenizers}

Our default tokenization strategy uses a Byte-Pair Encoder (BPE) from the HuggingFace tokenizers library \cite{huggingface}. BPE is a hybrid between character and word-level representations, which allows for the handling of large vocabularies in natural language corpora. Motivated by the intuition that rare and unknown words can often be decomposed into multiple known subwords, BPE finds the best word segmentation by iteratively and greedily merging frequent pairs of characters \cite{byte}. We compare this tokenization algorithm with a custom SmilesTokenizer based on a regex from \cite{schwaller2019molecular}, which we have released as part of DeepChem \cite{deepchem}.\footnote{\url{https://deepchem.readthedocs.io/en/latest/tokenizers.html\#smilestokenizer}}

To compare tokenizers, we pretrained two identical models on the PubChem-1M set. The pretrained models were evaluated on the Tox21 SR-p53 task. We found that the SmilesTokenizer narrowly outperformed BPE by $\Delta \textrm{PRC-AUC} = +0.015$. Though this result suggests that a more semantically-relevant tokenization may provide performance benefits, further benchmarking on additional datasets is needed to validate this finding.

\subsection{SMILES vs. SELFIES}
In addition to SMILES, we pretrained ChemBERTA on SELFIES (SELF-referencing Embedded Strings) \cite{selfies}. SELFIES is an alternate molecular string representation designed for machine learning. Because every valid SELFIES corresponds to a valid molecule, we hypothesized that SELFIES would lead to a more robust model. However, we found no significant difference in downstream performance on the Tox21 SR-p53 task. Further benchmarking is needed to validate this finding.

\subsection{Attention Visualization}

We used BertViz \cite{bertviz} to inspect the attention heads of ChemBERTa (SmilesTokenizer version) on Tox21, and contrast them to the molecular graph visualization of an attention-based GNN. We found certain neurons that were selective for chemically-relevant functional groups, and aromatic rings. We also observed other neurons that tracked bracket closures -- a finding in keeping with results on attention-based RNNs showing the ability to track nested parentheses \cite{suzgun2019lstm, yu2019learning}.

\begin{figure}[t!]
  \begin{subfigure}[b]{0.32\textwidth}
    \includegraphics[width=0.9\linewidth]{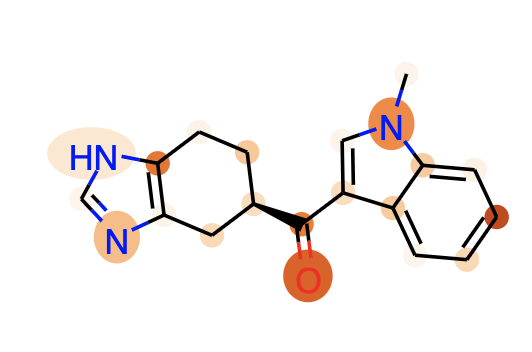}
    \caption{ }
    \label{fig:1}
  \end{subfigure}
  \begin{subfigure}[b]{0.32\textwidth}
    \includegraphics[width=0.9\linewidth]{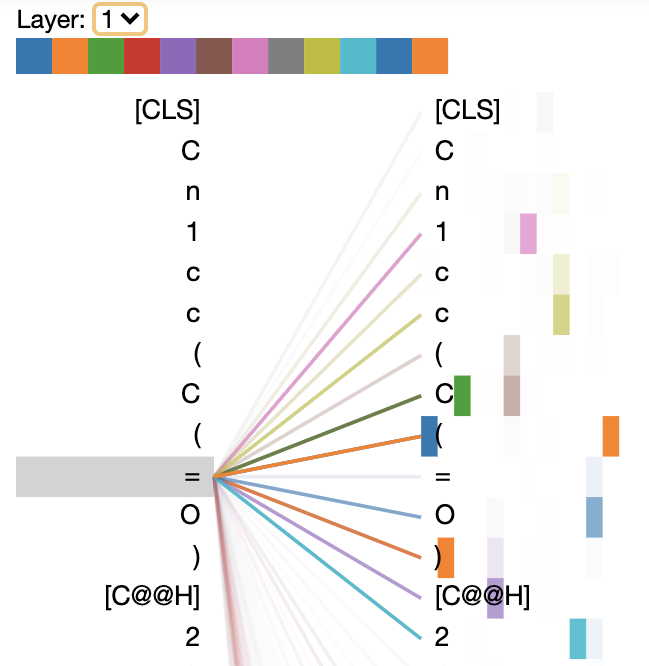}
    \caption{ }
    \label{fig:2}
  \end{subfigure}
  \begin{subfigure}[b]{0.32\textwidth}
    \includegraphics[width=0.9\linewidth]{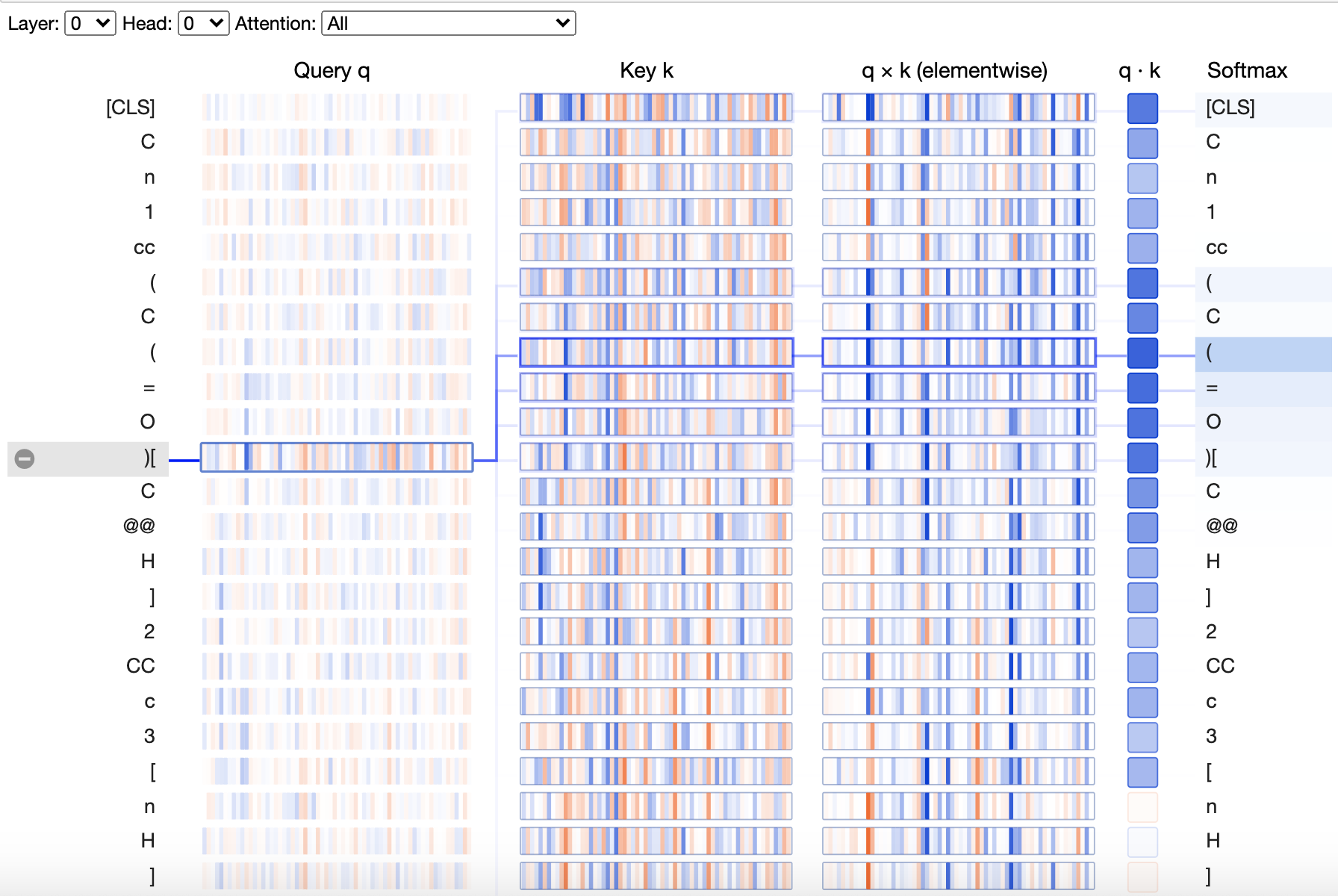}
    \caption{ }
    \label{fig:3}
  \end{subfigure}
  
  \caption{(a) Attention in GNNs highlights a problematic ketone in a Tox21 compound. (b) Attention over SMILES tokens in ChemBERTa provides a close analogue to graph attention. (c) Neural stack trace enables fine-grained introspection of neuron behavior. (b - c) produced via BertViz \cite{bertviz}.}
\end{figure}


\section{Discussion}
In this work, we introduce ChemBERTa, a transformer architecture for molecular property prediction. Initial results show that MLM pretraining provides a boost in predictive power for models on selected downstream tasks from MoleculeNet. However, with the possible exception of Tox21, ChemBERTa still performs below state-of-the-art on these tasks.

Our current analysis covers only a small portion of the hypothesis space we hope to explore. We plan to expand our evaluations to all of MoleculeNet, undertake more systematic hyperparameter tuning, experiment with larger masking rates, and explore multitask finetuning. In parallel, we aim to scale up pretraining, first to the full PubChem 77M dataset, then to even larger sets like ZINC-15 (with 270 million compounds). This work will require us to improve our engineering infrastructure considerably.

As we scale up, we are also actively investigating methods to improve sample efficiency. Alternative text-based pretraining methods like ELECTRA may be useful \cite{electra}. Separately, there is little question that graph representations provide useful inductive biases for learning molecular structures. Recent hybrid graph transformer models \cite{maziarka2020molecule, anonymous2021modelling} may provide better sample efficiency while retaining the scalability of attention-based architectures.

\section*{Broader Impact}

A core goal of AI for drug discovery is to accelerate the development of new and potentially life-saving medicines. Research to improve the accuracy and generalizability of molecular property prediction methods contributes directly to these aims. Nevertheless, machine learning---and particularly large-scale pretraining of the form we undertake here---is a resource-intensive process that has a growing carbon footprint \cite{lacoste2019quantifying}. According to the Machine Learning Emissions Calculator (\url{https://mlco2.github.io/impact}), we estimate that our pretraining generated roughly 17.1 kg CO\textsubscript{2}eq (carbon-dioxide equivalent) of emissions. Fortunately, Google Cloud Platform, which we used for this work, is certified carbon-neutral and offsets 100\% of emissions (\url{https://cloud.google.com/sustainability}). Even as we advocate for further exploration of large-scale pretraining for property prediction, we also encourage other researchers to be mindful of the environmental impact of these efforts and opt for sustainable cloud compute solutions where possible.

\begin{ack}

We would like to thank the Tyler Cowen and the Emergent Ventures fellowship for providing the research grant to S.C. for cloud computing and various research expenses, alongside the Thiel Foundation for funding the grant. Thanks to Mario Krenn, Alston Lo, Akshat Nigam, Professor Alan Aspuru-Guzik and the entire Aspuru-Guzik group for early discussions and mentorship regarding the potiential for applying large-scale transformers on molecular strings, as well as in motivating the utilization of SELFIES in this work. 

We would also like to thank the entire DeepChem team for their support and early discussions on fostering the ChemBERTa concept, and helping with designing and hosting the Tokenizers API and ChemBERTa tutorial. Thanks to the Reverie team for authorizing our usage of the PubChem 77M dataset, which was processed, filtered and split by them.
\end{ack}

\bibliography{main}
\bibliographystyle{unsrt}

\end{document}